\documentclass[letterpaper, 10 pt, conference]{ieeeconf}
\IEEEoverridecommandlockouts
\usepackage[utf8]{inputenc} 
\usepackage[T1]{fontenc}   
\usepackage{url}           
\usepackage{booktabs}       
\usepackage{amsfonts}      
\usepackage{nicefrac}      
\usepackage{microtype}      
\usepackage{xcolor}         
\usepackage{textcomp, gensymb} 
\usepackage{amsmath}       
\usepackage{csquotes}
\usepackage{float}
\usepackage{caption}
\usepackage{graphicx}      
\usepackage{subcaption}
\usepackage{algorithm}
\usepackage{algpseudocode}
\usepackage{amssymb}
\usepackage{multirow}
\usepackage{caption}
\usepackage[hidelinks]{hyperref}

\DeclareMathOperator*{\argmin}{argmin}

\title{\LARGE \bf
$IR^2$: Implicit Rendezvous for Robotic Exploration Teams under 

Sparse Intermittent Connectivity
}

\author{Derek Ming Siang Tan$^{1,2}$$^{\dagger}$, Yixiao Ma$^{1}$, Jingsong Liang$^{1}$, Yi Cheng Chng$^{2}$, Yuhong Cao$^{1}$, Guillaume Sartoretti$^{1}$
\thanks{$\dagger$ Corresponding author, to whom correspondence should be addressed.}
\thanks{$^{1}$Authors are with the Department of Mechanical Engineering, College of Design and Engineering, National University of Singapore 
{\tt\small \{derektan,yixiaoma,jingsongliang\}@u.nus.edu,
\{caoyuhong, mpegas\}@nus.edu.sg}
}
\thanks{$^{2}$Author is with Singapore Technologies Engineering Ltd,
        {\tt\small yicheng.chng@stengg.com}}
}

\begin{document}
\maketitle
\thispagestyle{empty}
\pagestyle{empty}

\begin{abstract}

Information sharing is critical in time-sensitive and realistic multi-robot exploration, especially for smaller robotic teams in large-scale environments where connectivity may be sparse and intermittent. Existing methods often overlook such communication constraints by assuming unrealistic global connectivity. Other works account for communication constraints  (by maintaining close proximity or line of sight during information exchange), but are often inefficient. 
For instance, preplanned rendezvous approaches typically involve unnecessary detours resulting from poorly timed rendezvous, while pursuit-based approaches often result in short-sighted decisions due to their greedy nature. 
We present \textbf{$IR^2$}, a deep reinforcement learning approach to information sharing for multi-robot exploration. 
Leveraging attention-based neural networks trained via reinforcement and curriculum learning, \textbf{$IR^2$} allows robots to effectively reason about the longer-term trade-offs between disconnecting for solo exploration and reconnecting for information sharing.
In addition, we propose a hierarchical graph formulation to maintain a sparse yet informative graph, enabling our approach to scale to large-scale environments.
We present simulation results in three large-scale Gazebo environments, which show that our approach yields $6.6 - 34.1\%$ shorter exploration paths when compared to state-of-the-art baselines, and lastly deploy our learned policy on hardware.
Our simulation training and testing code is available at~\textcolor{magenta}{\url{https://ir2-explore.github.io}}.

\end{abstract}

\section{INTRODUCTION}

Multi-robot exploration of unknown environments is a well-established research area, with significant improvements in exploration planning speed and scalability of robot team sizes in recent years. It also has a large and growing number of applications such as underwater and planetary exploration~\cite{underwater_explore,planet_explore1}, underground mining~\cite{underground_mining}, and search and rescue~\cite{search_rescue2}. However, one main challenge of translating multi-robot exploration research into real-world applications is accounting for realistic inter-robot communication constraints. 

In the real world, information exchange can only occur when robots are connected. Such connectivity between robots is often limited by signal strength between communication devices, which can be modeled as a function of distance and the medium in which communication is carried out~\cite{rssi_model}. Some exploration planners assume unrealistic global connectivity, where robots remain continuously connected regardless of the changing distance between them or the communication medium~\cite{smmr_explore}. This assumption does not hold in reality. Meanwhile, other exploration planners adopt different information-sharing strategies to account for communication constraints. By doing so, these planners can achieve more effective generalization to real-world scenarios. 

\begin{figure}
    \centering
    \includegraphics[width=0.95\linewidth]{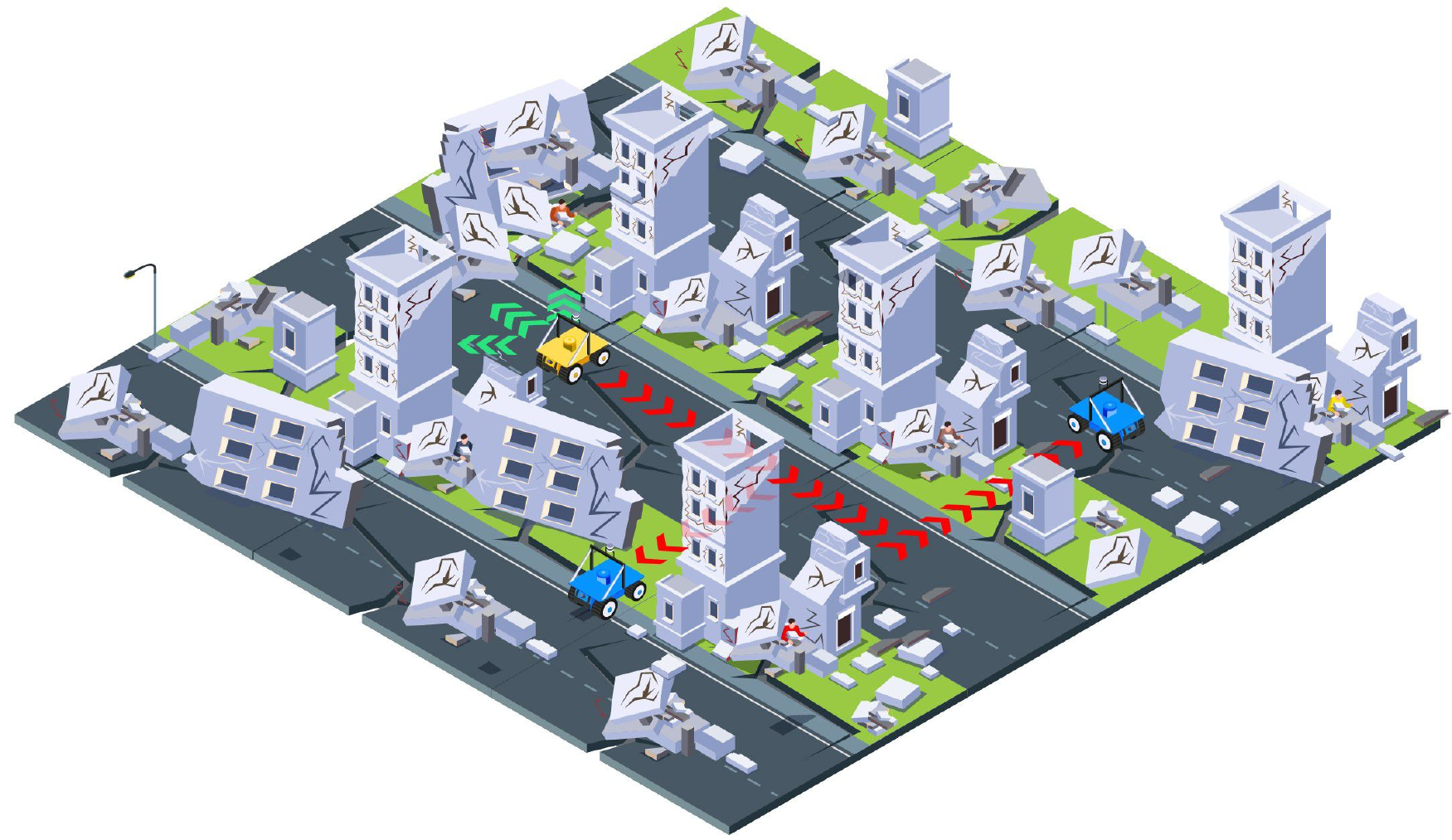}
    \caption{Search-and-rescue robots collaborate to navigate uncharted disaster zones under realistic communication constraints. For example, the yellow robot must balance trade-offs between disconnecting from its team (blue robots) to independently explore (green arrows), and pursuing other robots to exchange information (red arrows).}
    \label{fig:enter-label}
    \vspace{-20pt} 
    \end{figure}
    
Connectivity strategies in multi-robot exploration can be divided into three broad categories. The first category, \textit{opportunistic connectivity,} involves robots focusing on exploration while leaving connectivity to chance. This may lead to poor decisions due to incomplete environmental knowledge~\cite{opp_conn1}~\cite{gvg_explore}. The second category, \textit{continuous connectivity}, requires robots to maintain continuous connection during exploration, potentially sacrificing efficiency~\cite{base_station_explore}~\cite{acord_explore}. The third category, \textit{recurrent connectivity}, permits robots to actively connect and disconnect, allowing for more efficient information sharing and exploration \cite{lauren_explore}~\cite{hkust_explore}. Under \textit{recurrent connectivity}, a common approach is for all robots to establish explicit consensus on when and where to meet (i.e., a \textit{rendezvous} point) before breaking off to explore the environment independently. While this guarantees frequent information sharing, exploration efficiency is sacrificed when robots are forced to backtrack along long paths with minimal information gain for poorly-timed rendezvous. Another common approach are the pursuit-based methods~\cite{mtare_planner}, which avoids the use of explicit consensus by pursuing other robots to share information when the benefits outweigh the cost in terms of exploration efficiency. 
However, such approaches are often short-sighted and greedy based on their current world belief without considering future implications.

To this end, we propose a novel information-sharing strategy that achieves high exploration efficiency by estimating the future impact of current exploration and rendezvous decisions. 
There are three key contributions to our proposed work.
First, we use an \textit{attention-based neural network} trained by deep reinforcement learning (DRL) to help robots learn to sequence non-myopic decisions. Second, we implement \textit{two-stage curriculum learning}, where robots are placed in increasingly difficult exploration environments with increasing frequency and duration of disconnectivity. This drives robots to learn complex, dynamic connectivity strategies to attain even higher exploration efficiency. Lastly, we utilize a \textit{hierarchical graph formulation}, to enable scaling of our strategy to large-scale environments. This involves maintaining both a sparse global graph representation of the robots' map and a dense local graph centered on the robot. Combining graphs at different spatial scales helps robots strike a balance between long- and short-term exploration and rendezvous goals.

Based on simulations, our method outperforms state-of-the-art preplanned and pursuit-based planner
baselines by $6.6 - 34.1\%$ in distance efficiency while significantly improving mapped area consistency among robots. This indicates strong collaboration to achieve more equal and effective sharing of the exploration task. We believe this is because our method enables robots to effectively balance the longer-term trade-offs between disconnecting for solo exploration and reconnecting for information sharing. As the deployment of large-scale robotic systems remains prohibitive in terms of financial cost and hardware complexities, our paper focuses on relatively small but highly effective robotic exploration teams~\cite{scalability_challenges}.

\section{RELATED WORK}

\subsection{Multi-Robot Exploration}

\subsubsection{\textbf{Conventional Planners}}

There is extensive literature on multi-robot exploration using conventional planners. Existing planners can be frontier-based, trajectory optimization-based, or potential-based. Kulkarni et al. \cite{gb_planner} proposed a 3D rapidly random graph sampling method to plan paths for both UAVs and UGVs through a centralized planner. Cao et al.~\cite{mtare_planner} divided the exploration space into sub-volumes and distributed them to multiple robots via a centralized planner by treating the problem as a vehicle routing problem. Yu et al.~\cite{smmr_explore} proposed a decentralized approach that uses artificial potential fields to attract multiple robots to different frontiers, and to repulse them away from one another. Although conventional planners are reliable, they often rely on greedy strategies to plan short-term paths due to their inability to reason about the future impacts of their current decisions.

\subsubsection{\textbf{Learning-based Planners}}

In the recent decade, DRL-based planners have shown remarkable performance due to their ability to estimate and maximize long-terms returns. Yu et al.~\cite{async_mappo} used asynchronous multi-robot proximal policy optimization to train robots to efficiently explore unknown environments, while using action-time randomization to effectively transfer simulation to real-world experiments. Luo et al.~\cite{gnn_explore} utilized graph convolutional neural networks to achieve efficient multi-robot exploration. Cao et al.~\cite{ariadne_explore} relied on attention-based neural networks to achieve long-term planning in exploration, albeit for a single robot. In this paper, we extend our previous work~\cite{ariadne_explore} to multi-robot exploration while considering communication constraints.

\subsection{Connectivity Strategies}

\subsubsection{\textbf{Opportunistic Connectivity}}

Some exploration methods defined connectivity to occur by chance~\cite{opp_conn1}~\cite{gvg_explore}. As encounters are random, such approaches do not offer completion guarantees and often have high performance variance.

\subsubsection{\textbf{Continuous Connectivity}}

Many works, including ~\cite{base_station_explore}~\cite{acord_explore}, achieved exploration while ensuring robots were continuously connected to each other as well as to the base station. While these methods ensure a consistent understanding of the environment and allow for centralized planning, they sacrifice efficiency as robots are often not well-distributed.

\subsubsection{\textbf{Recurrent Connectivity}}

Preplanned approaches ~\cite{lauren_explore}\cite{hkust_explore} advocated for explicit consensus among robots in deciding where and when to gather during frontier-based exploration. While these approaches guarantee frequent information sharing, some exploration efficiency is sacrificed when robots are forced to backtrack along paths with minimal information gain resulting from poorly timed rendezvous. On the other hand, pursuit-based approaches~\cite{mtare_planner} pursue other robots to share information when the estimated benefits outweigh the cost in terms of exploration efficiency. However, these approaches often remain short-sighted, and tend to act greedily based on their current world belief. In this paper, we leverage attention-based neural network trained with reinforcement and curriculum learning to enhance existing information-sharing strategies with the ability to estimate the future impact of current actions.

\begin{figure*}[tb]
    \centering
    \includegraphics[width=1.0\linewidth]{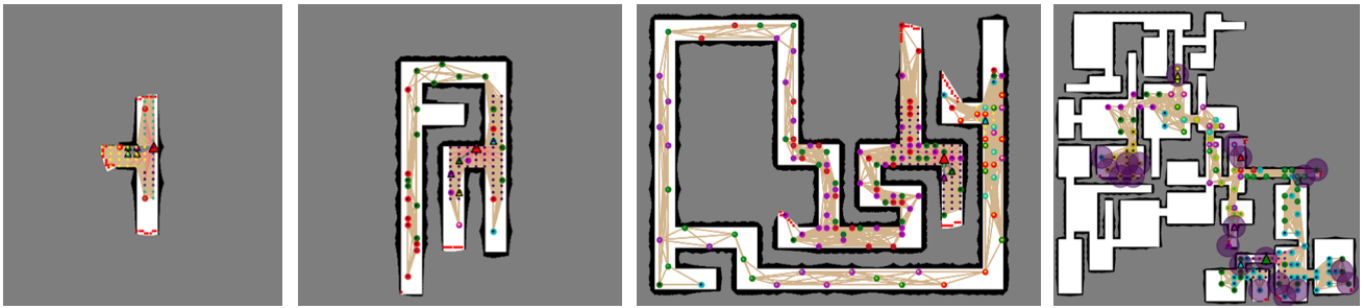}
    \vspace{-0.6cm}
    \caption{\textbf{Hierarchical Graph Formulation} Four-stage process illustrated with snapshots from different episodes: (a) Dense \textit{local graph construction} around robot's position. (b) Sparse \textit{global graph construction} via offshoots toward frontiers. (c) \textit{Global graph merger} combining different robots' global graphs (different colored nodes). The map-surplus utility paths ($s_{i,j}$) between robots are shown as paths with colors of increasing intensity (black to yellow). (d) \textit{Global graph pruning} to remove nodes that do not lead to frontiers centers (purple circles).}
    \label{fig:graph_formulation}
    \vspace{-10pt} 
\end{figure*}

\section{PROBLEM FORMULATION}
\label{problem_formulation}

We extend the single-robot exploration problem formulation in~\cite{ariadne_explore} into a multi-robot exploration problem with connectivity constraints. We consider a bounded and unknown environment represented by a 2D occupancy grid map denoted by $\mathcal{M}$. We have $n$ robots $\{1,2,...,n\}$, and each robot maintains its own belief over the exploration map $\mathcal{M}_i$. Each map belief consists of unknown region $\mathcal{M}_u$ and known region $\mathcal{M}_k$, where $\mathcal{M}_u \cup \mathcal{M}_k = \mathcal{M}$. 
$\mathcal{M}_k$ can be further divided into free area $\mathcal{M}_f$ and occupied $\mathcal{M}_o$, such that $\mathcal{M}_f \cup \mathcal{M}_o = \mathcal{M}_k$.
Each robot is equipped with a 360$\degree$ lidar, with sensor range $d_s$. At the beginning of each exploration mission, we assume that the relative positions of all robots are known.  

We define the trajectory of viewpoints for each robot $\psi_i = (\psi_{i,1}, \psi_{i,2},\ldots,\psi_{i,m})$, $\psi_{i,j} \in \mathcal{M}_i$. This setup presents an optimization problem where we seek an optimal trajectory set $\Psi^*$ given all possible trajectory sets $\Psi$, which minimizes the maximum of all robot's trajectory length $L(\psi_i)$. The goal $\mathcal{M}_g$ is for all robots to achieve 99\% exploration of the ground truth free space in their individual map belief. 

\vspace{-0.4cm}
\begin{equation} \label{eq2}
    \begin{aligned}
    \Psi^* = \argmin_{\psi \in \Psi}\max_{i \in [1,n]}( L(\psi_i) ) 
    \text{, \hspace{0.25cm} s.t. $\forall{\mathcal{M}}_i = \mathcal{M}_g$ } 
    \end{aligned}
\end{equation}

Robots exchange map, graph, and position information whenever they are within communication range. Their individual exploration map beliefs are merged to form $\mathcal{M}_k = \bigcup_{i \in [k]}\mathcal{M}_{k,i}$. Proximity and signal strength are frequently used in the research community to define connectivity. For proximity-based communication, robots are considered to be connected when they are within a specified communication distance. Alternatively, communication range based on signal strength can be defined using the log distance path loss (LDPL) model~\cite{rssi_model}. The LDPL model predicts the signal strength attenuation that a signal encounters when propagated through different types of environments. Such an attenuation is also known as path loss, $PL = P_T - P_R$, where $P_T$ and $P_R$ are the transmitted and received power respectively. Any two robots are considered to be connected when $P_R >= P_{thresh}$. To factor in the effects of obstacles on signal strength, we adopt the modified LDPL model formulated by~\cite{flocking_obstacle_comms}, and chose parameters that closely matches the realistic model~\cite{rssi_model}.

\section{EXPLORATION AS AN RL PROBLEM}

\begin{figure*}[tb]
    \centering
    \includegraphics[width=1.0\linewidth]{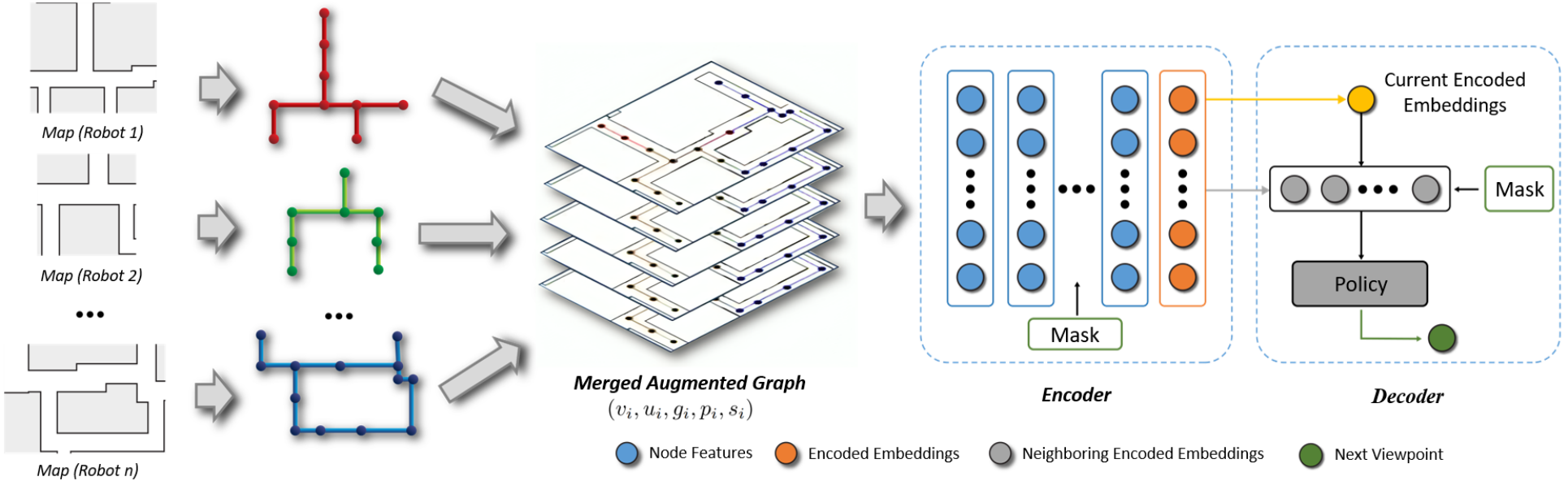}
    \vspace{-0.45cm}
    \caption{\textbf{DRL-Based Planner Architecture.} In a multi-robot setting, for any robot, our approach first merges and sparsifies the global graphs shared by robots within connectivity range. Thereafter, we augment the merged graph with additional information pertinent to exploration and rendezvous. This augmented graph is fed into a similar encoder-decoder attention-based neural network architecture as~\cite{ariadne_explore}.}
    \label{fig:model}
    \vspace{-10pt}
\end{figure*}

We formulate the multi-robot exploration problem under communication constraints as a decentralized partially observable Markov Decision Process (Dec-POMDP)~\cite{dec_pomdp}, represented by the tuple $(\mathcal{N}, \mathcal{S}, \mathcal{A}, \mathcal{R}, \mathcal{T}, \mathcal{O}, \mathcal{Z}, \mathcal{\rho}, \mathcal{\gamma})$ with $\mathcal{N}$ robots, state space $\mathcal{S}$, action space $\mathcal{A}$, reward function $\mathcal{R}$, state transition function $\mathcal{T}$, observation set $\mathcal{O}$ sampled from observation functions $\mathcal{Z}$, initial states $\mathcal{\rho}$, and discount factor $\mathcal{\gamma}$. At each decision step, each robot selects and take an action based on its policy $\pi(a^t_i|o^t_i)$. Each robot $i$ aims to maximize its own total expected return $ R_i = \mathbb{E}_{a_t \sim \pi(\cdot \mid o_t)}\left[\sum_{t=1}^{T} \gamma^{t-1} r^t_i \right].$

\subsection{Hierarchical Graph Formulation}

To avoid overwhelming each robot with dense grid-based map data, we extract from their map belief $\mathcal{M}_i$ a hierarchical collision-free graph $G^t_i = (V^t_i, E^t_i)$, where $V_i = (v_{i,1}, v_{i,2}, \ldots, v_{i,m})$, $\quad \forall v_{i,j} = (x_{i,j}, y_{i,j}) \in \mathcal{M}_f$ represent candidate viewpoints that are distributed across the free space of the map. Inspired by~\cite{mtare_planner}, we maintain a global sparse graph representation $G^t_{i,S} = (V^t_{i,S}, E^t_{i,S})$ of the map and a local dense graph $G^t_{i,D} = (V^t_{i,D}, E^t_{i,D})$ around the robot. We observe that having two graphs at different spatial scales will help robots balance between long- and short-term exploration and rendezvous goals. Unlike previous work that performed path planning on the graphs separately~\cite{mtare_planner}~\cite{gb_planner}, we combine both graph representations into a single graph for path planning, where $V^t_i = \{ V^t_{i,S} \cup V^t_{i,D} \}$. Thereafter, we generate collision-free edges $E^t_i$ by connecting each vertex $v^t_{i,j}$ with its $k$ nearest neighbors that are within line of sight.

In large-scale environments, robots may experience poor long-term planning and slow planning rates due to the large number of vertices and edges in their collision-free graphs. To enhance exploration efficiency, we introduce four key features to generate a dense local graph and to retain a sparse global graph throughout the exploration process.

\subsubsection{\textbf{Local Graph Construction}}

At every graph update step, a set of local graph vertices surrounding the robot is extracted from vertices $V^t_i$,
where $v_{i,R}$ is the robot's position, and $d_r = 2d_s$ is the length of the box centered on $v_{i,R}$.
\begin{equation}
    \footnotesize
    V^t_{i,D} = \left\{ v \in V^t_i \ \middle| \ |v_{i,x} - v_{i,R,x}| \leq d_r \ \text{\&} \ |v_{i,y} - v_{i,R,y} | \leq d_r \right\}
\end{equation}

Thereafter, we generate collision-free edges $E_{t,D}$ by connecting each vertex with its $k$ nearest neighbors that are within line of sight. For each of these vertex, we calculate their exploration utility $u_i = (u^1_i, u^2_i, \ldots, u^m_i)$, which represents the number of observable frontiers which are line of sight from their respective positions.

\subsubsection{\textbf{Global Graph Construction}}

Inspired by the concept of \textit{rapidly random graphs}~\cite{gb_planner}, our \textit{global graph construction} process (Fig.~\ref{fig:graph_formulation}b) is ego-centric and comprises two parts. First, at every graph update step, we add the robot's position directly to its own global graph as it is guaranteed to be traversable by the robot. Second, we extend the global graph from the robot's position towards frontier clusters present in the dense local graph. Such clusters, also known as frontier centers, are defined as clusters of nonzero utility viewpoints separated by a threshold radius $r_g$. We use $A^*$ for graph extension, where the nodes along these paths are directly added to the global graph. Note that our approach is nonrandom, unlike the sampling method in \textit{rapidly random graphs}.

\subsubsection{\textbf{Global Graph Merger}}

The \textit{graph merger} algorithm (Fig.~\ref{fig:graph_formulation}c) is responsible for combining new global graph nodes from other robots, and for sparsifying the combined graph to maintain a minimal set of nodes for computational tractability. Whenever robots are connected, they share their global graphs with each other. We first perform Euclidean down-sampling of the incoming global graph nodes, and connect these new nodes to the current global graph using the same $k$ nearest-neighbor approach in the \textit{global graph construction} step. We then iterate through each global graph node to attempt to combine their neighboring nodes within a specified radius $r_m$. Here, we remove neighboring nodes if the entire graph remains connected when they are removed. Note that this approach relies on global graphs already constructed by other robots during their exploration, hence saving computation from not having to reconstruct graphs representing the new map portions recently merged in. We run this algorithm at every graph update step as it is computationally light.

\subsubsection{\textbf{Global Graph Pruning}}

The \textit{global graph pruning} algorithm (Fig.~\ref{fig:graph_formulation}d), inspired by~\cite{cao2024deep}, removes irrelevant branches, while maintaining connectivity between all robots and frontier clusters. To adapt the algorithm for a multi-robot setting, we build multiple Dijkstra cost graphs beginning from each robot's position, to obtain shortest paths between each robot and all frontier centers. We then reconstruct edges from these shortest path nodes to form the pruned graph. 
Compared to~\cite{cao2024deep}, our approach is more computationally efficient in a multi-robot setting, since the number of path planning runs scales linearly with the number of robots. This algorithm is invoked every $N_p$ iterations.

\subsection{Observation Space}
\label{subsec:obv}
The observation for each robot $i$ is $o_t=G^{'t}_i$, where $G^{'t}_i = (V^{'t}_i, E^{'t}_i)$ is the augmented graph modified from $G_t$. Each augmented vertex $v'_{i,j} = (v_{i,j}, u_{i,j}, g_{i,j}, p_{i,j}, s_{i,j})$ comprises five components, of which the first three components belong to the original single-robot formulation~\cite{ariadne_explore}. The exploration utility $u_{i,j}$ represents the number of observable frontiers within the node $v_{i,j}$'s line of sight. The guidepost $g_{i,j}$ is a binary value indicating if a node has been visited by robot $i$.

We further augment each node with two additional features to allow robots to better cooperate in the exploration task, and to decide whether to pursue other robots for information sharing. First, the position indicator $p_{i,j}$ denotes whether a node is occupied by the current robot, by another robot, or unoccupied, with values -1, +1, and 0 respectively. 
This indicates the relative location of other robots with respect to the current robot, allowing for better cooperative decisions to be made. Note that $p_{i,j}$ is updated only when robots $i$ and other robots are within communication range.

Second, the map-surplus utility $s_{i,j}$ indicates how much additional map information a robot believes it possesses relative to other robots. This is represented by $A^*$ paths along the hierarchical graph that connects the current robot $i$ to other robots. The map-surplus utility values corresponding to each of these path nodes increases linearly as the distance to the other robots gets smaller.

\vspace{-0.25cm}
\begin{equation}
    \label{map_surplus_observation}
    \small
    s_{i,j} = 
    \begin{cases} 
    \begin{aligned}
    &d_{i,j} \frac{\Delta{M_{i,k}} - s_{\text{min}}}{d_{i,k}} + s_{\text{min}} \\
    \end{aligned}
    \vspace{-0.5cm}
    & \text{if } \Delta{M_{i,k}} \geq \Delta{M_{\text{min}}} \\[20pt]
    0, & \text{otherwise}
    \end{cases}
\end{equation}

where $d_{i,j}$ represents the distance between node $v_{i,j}$ and robot $i$'s position along the $A^*$ path, $\Delta{M_{i,k}} = M_{i,i} - M_{i,k}$ the map area difference perceived by robot $i$ relative to robot $k \in \{1,2,...,n\}$, $\Delta{M_{\text{min}}}$ the minimum map area difference to consider a non-zero $s_{i,j}$, and $s_{\text{min}}$ a minimum constant value. Intuitively, the map-surplus utility guides robots towards other robots via a path of increasing utility corresponding to the perceived map surplus. 
In cases where there are multiple overlapping $A^*$ paths leading to different robots, the map-surplus utility of the overlapping node is determined by the highest utility value among all paths at that location.

\subsection{Action Space} 

The action space consists of the $k$ nearest neighboring graph vertices that are in the robot's line of sight. Given observation $o_t$ at every decision step, each robot's attention-based neural network outputs a stochastic policy, denoted as $\pi_{\theta,i}(a^t_i|o^t_i) = \pi_{\theta,i}(\psi^{t+1}_{i,j}=v_{i,j},(\psi^t_{i,j}, v_{i,j})\in E^t_i \mid o^t_i)$. We sample each robot's action following a multinomial distribution during training, and greedily during inference.

\subsection{Reward Structure}

Our main objective is for robots to make long-term decisions that balance trade-offs between individual exploration and pursuing other robots for information sharing. We retain the three reward components introduced in the single-robot formulation~\cite{ariadne_explore}. This includes the reward for the number of observable frontiers from the new viewpoint $r_o = | F_{o, \psi^{t+1}_i} |$, the distance penalty between the current and new viewpoint $r_d = -C(\psi^t_i, \psi^{t+1}_i)$, and exploration completion reward $r_c$. 

To achieve better coordination and timely information sharing among multiple robots, we introduce two additional reward components. The first $r_f = \Delta ( F_{\psi^{t}_i}, F_{\psi^{t+1}_i} ) $ refers to the increase in the total number of frontier points on the combined map assuming no communication constraints $\mathcal{M}^t = (\mathcal{M}_1 \cup \mathcal{M}_2 \ldots \mathcal{M}_n)$. Note that the robots do not directly observe this combined map. Instead, it is only used as privileged information during training to avoid providing incentives for redundant exploration that does not contribute to the team's overall exploration task. The second $r_s = | s_{i,\psi^{t+1}_i} | $ refers to the incentive of being on a position along the path of map-surplus utility that robots can observe. This provides dense rewards that encourage robots to pursue other robots for information sharing. The total reward is computed as $\begin{aligned}   
\label{eq:reward}
r_t(o_t, a_t) = \alpha_1 \cdot r_o + \alpha_2 \cdot r_d + \alpha_3 \cdot r_f + \alpha_4 \cdot r_s  + r_c.
\end{aligned}$

\section{NEURAL NETWORK AND TRAINING}

\subsection{Policy and Critic Networks}
\label{subsec:network}

\subsubsection{\textbf{Encoder}}

The encoder transforms the explored map into a multi-scale representation through multiple self-attention layers~\cite{attention}. First, we embed augmented graph vertices $V'_i$ into d-dimension feature vector $h_i^n$ . As seen in Eq.~\eqref{Attention equation}, we obtain the query, key, and value vectors $q_i$, $k_i$, and $v_i$ by multiplying their feature vectors $h_i^q = h_i^k = h_i^v =h_i^n$ with their learnable weight matrices $W^Q$, $W^K$, and $W^V$ respectively. We then compute the similarity matrix $u_{ij}$ and the attention weights $w_{ij}$ to obtain the enhanced node feature $h'_i$. Note that each attention layer takes the output of the previous one as input. 

\vspace{-0.3cm}
\begin{equation}
    \label{Attention equation}
    \begin{aligned}
    q_i &= W^Qh_i^q, & k_i &= W^Kh_i^k, & v_i &= W^Vh_i^v \\
    u_{ij} &= \frac{q_i^T \cdot k_j}{\sqrt{d}}, & w_{ij} &= \frac{e^{u_{ij}}}{\sum_{j=1}^n e^{u_{ij}}}, & h_i^\prime &= \sum\limits_{j=1}^n w_{ij}v_j
    \end{aligned}
\end{equation}

\subsubsection{\textbf{Decoder}}

The decoder outputs a policy for the robot to act upon. We first extract the current node features $h^c$ based on the current robot position and its connected neighboring node features $h^n$ from the output of the encoder $h'_i$. Thereafter, we pass these features into a cross-attention layer, where $h^q = h^c$, $h^k = h^n$, and $h^v = h^n$. Similar to Eq.~\eqref{Attention equation}, we eventually obtain the output feature vector. It is then concatenated with $h^c$ and projected back into a d-dimension enhanced current node feature vector $\hat{h}^c$.  Finally, we pass both $\hat{h}^c$ and $h^{n}$ into a pointer layer~\cite{pointer} to output the robot policy.

\begin{figure}
    \centering
    \includegraphics[width=1.0\linewidth]{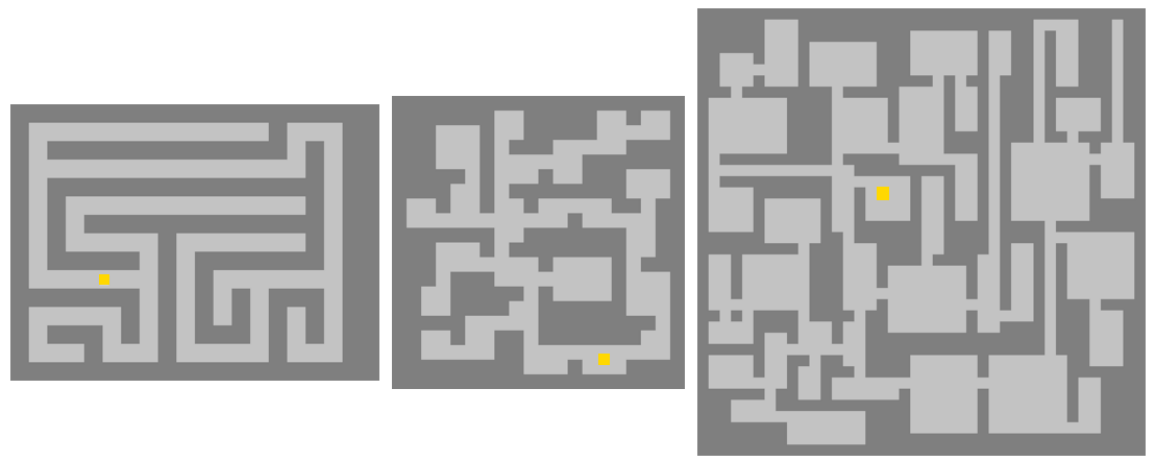}
    \vspace{-0.6cm}
    \caption{\textit{Corridor}, \textit{Hybrid}, and \textit{Complex} maps (left to right).}
    \label{fig:training_maps}
    \vspace{-15pt}
\end{figure}

\subsection{Training}

\begin{figure*}[t]
    \centering
    \includegraphics[width=1.0\linewidth]{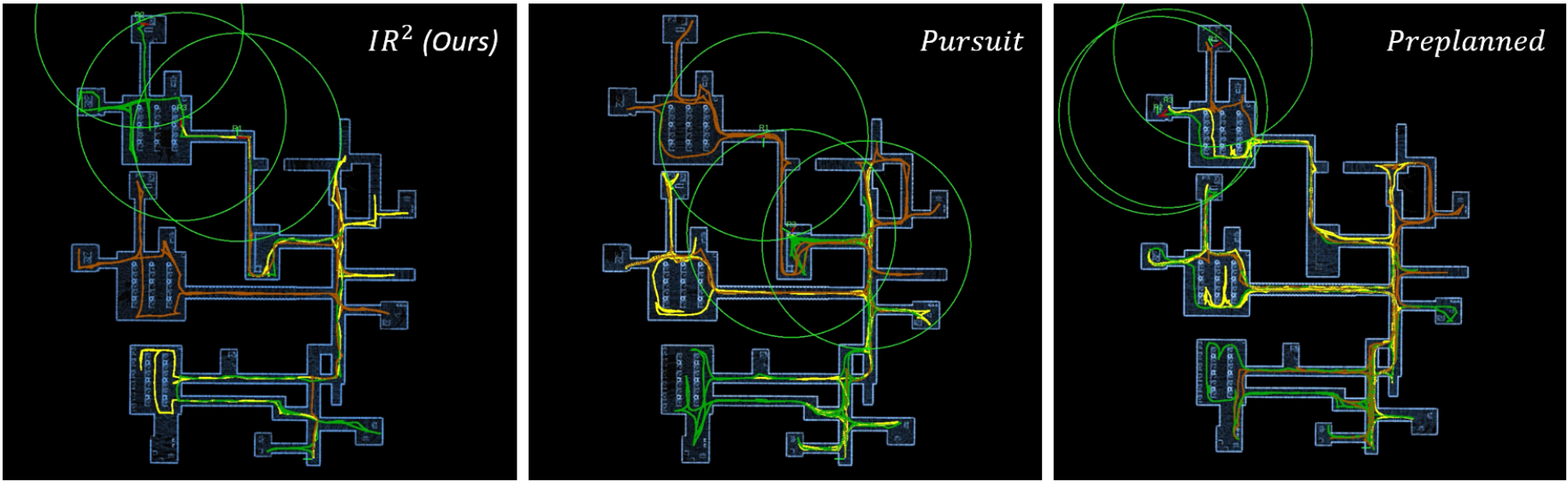}
    \vspace{-0.6cm}
    \caption{\textbf{Path visualization} of competing approaches for three robots in \textit{Indoor}~\cite{mtare_planner}, where there is no opportunistic connectivity across walls. We observe $IR^2$ exhibits the least trajectory overlap and backtracking, followed by \textit{Pursuit}~\cite{mtare_planner}, and finally \textit{Preplanned}~\cite{lauren_explore}.}
    \label{fig:model}
    \vspace{-0.5cm}
\end{figure*}

\subsubsection{\textbf{Curriculum Learning}}

We develop a curriculum~\cite{curriculum} to allow robots to learn complex information-sharing strategies incrementally. 
Our training data contains four types of maps, namely \textit{Simple}, \textit{Corridor} ($\text{160}m \times \text{120}m$), \textit{Hybrid} ($\text{125}m \times \text{125}m$), and \textit{Complex} maps ($\text{250}m \times \text{250}m$). 
We obtain \textit{Simple} maps from an open-source data set~\cite{maps_open_source}, and custom-generated datasets for the remaining maps (Fig.~\ref{fig:training_maps}). 

Given these maps, we develop a curriculum to train robots in two stages - the first with an \textit{easy training set}, and the second with a \textit{difficult training set}. 
The \textit{easy training set} contains 10000 maps, with 5000 \textit{Simple} and \textit{Corridor} maps each. The \textit{Simple} maps train robots on basic exploration skills such as moving to frontiers efficiently, while the \textit{Corridor} maps train robots to handle situations with prolonged disconnectivity. 
The \textit{difficult training set} contains 6000 maps, with 2000 \textit{Corridor}, \textit{Hybrid}, and \textit{Complex} maps each. We introduce \textit{Hybrid} maps that pose challenges found in simple and corridor maps, to test robots on both sets of skills within the same environment. In addition, we introduce \textit{Complex} maps as a significantly tougher version of \textit{Hybrid} maps.

\subsubsection{\textbf{Training Details}}
\label{training_details}

We employ Ray~\cite{ray_training} to perform 32 concurrent training simulations, each with 3-5 robots for the \textit{easy training set} and 4-6 robots for the \textit{difficult training set}. We train our attention-based neural network using the soft actor-critic (SAC) algorithm~\cite{soft_actor_critic}, utilizing an AMD Ryzen threadripper 3970x and four NVIDIA A5000 GPUs. The task is considered successful only when all robots achieve 99\% exploration in their respective belief map. Each training run on the first training set takes approximately 6000 episodes and 18 hours to complete, while each training run on the second training set takes approximately 7500 episodes and 60 hours to complete.

\section{EXPERIMENTS}

\subsection{Experimental Setup}

\begin{figure}
    \centering
    \includegraphics[width=1.0\linewidth]{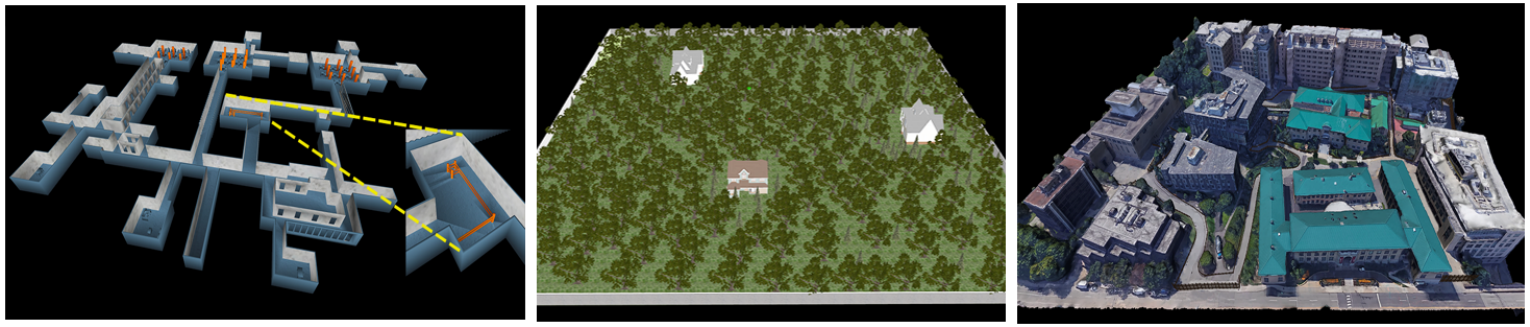}
    \vspace{-0.6cm}
    \caption{\textit{Indoor}, \textit{Forest}, and \textit{Campus} Gazebo maps (left to right)~\cite{mtare_planner}.}
    \label{fig:training_maps}
    \vspace{-18pt}
\end{figure}

We utilize an open-source multi-robot exploration benchmark~\cite{mtare_planner} to validate the performance and generalizability of our trained model. We test our approach in three large-scale Gazebo simulation environments - \textit{Indoor} ($\text{130}m \times \text{100}m$), \textit{Forest} ($\text{150}m \times \text{150}m$), and \textit{Campus} ($\text{340}m \times \text{340}m$). We blocked off access to overlapping pathways in the \textit{Campus} environment to ensure compatibility with our 2D planner. 

In addition, these simulation environments offers realistic robot kinematic and sensor models compared to our simplified training setup. Each ground vehicle has a preset maximum speed of 2 $m/s$ and is equipped with a Velodyne Puck lidar for exploration and mapping.  We utilize Octomap~\cite{octomap} to generate 2D occupancy grid maps from 3D lidar scans, with a mapping range of $20m$. We perform experiments for both proximity- and signal strength-based communication models (defined in Sec. \ref{problem_formulation}). Whenever two robots can communicate, we combine their maps using a modified 2D Map-Merge package~\cite{explore_bench} and update their knowledge of each other's positions and global graphs. We enable information hopping through intermediate robots to make information sharing more realistic. Similar to training, we terminate each run after all robots have explored 99\% of the ground truth map.  All experiments are conducted using the same training resources to ensure repeatability (Sec.~\ref{training_details}).

\subsection{Comparison and Analysis}

\begin{table*}[!ht]
  \vspace{2mm}
  \centering
  \caption{Experimental results (30m proximity connectivity). The notation '$\uparrow$' implies that a larger value is preferable, and vice versa. 
  Values in parentheses next to each data point denote the corresponding standard deviation. }
  \vspace{-2mm}
  \scalebox{0.71}{
    \begin{tabular}{l||ccc||ccc||ccc||ccc}
    \toprule
    \textbf{Model} & 
    \multicolumn{3}{c}{\textbf{Time Efficiency, $\eta_T$ ($m^3/s$)} $\uparrow$} &
    \multicolumn{3}{c}{\textbf{Distance Efficiency, $\eta_D$ ($m^3/m$)} $\uparrow$} &
    \multicolumn{3}{c}{\textbf{Map Area Stdev, $\sigma$ (\%)} $\downarrow$} &
    \multicolumn{3}{c}{\textbf{Computing Time ($s$)} - Planning / Total Time $\downarrow$}\\
    \midrule
    \cmidrule{2-13}

    & \multicolumn{12}{c}{\textbf{130m $\times$ 100m Indoor environment with 2, 3, 4 robots}} \\
    
    \midrule\midrule    
    Preplanned ~\cite{lauren_explore}   
                  & 5.8 (0.7)   & 8.1 (0.3)    & 8.1 (0.6)
                  & 4.5 (0.4)   & 5.8 (0.1)    & 5.9 (0.2) 
                  & 4.2 (2.2)   & \textbf{4.6 (0.9)}    & \textbf{4.3 (0.8)} 
                  & 0.14 / \textbf{0.14 (0.09)}  & 0.12 / \textbf{0.12 (0.04)} & 0.16 / \textbf{0.16 (0.06)}   \\ 
    Pursuit ~\cite{mtare_planner}    
               & 11.6 (1.2)    & \textbf{13.3 (5.2)}   & 13.5 (2.8)     
               & 6.6 (0.6)     & 8.1 (2.9)    & 8.5 (1.7)    
               & \textbf{2.2 (2.1)}     & 8.1 (4.1)    & 5.5 (4.8)    
               & 0.19 / 0.42 (0.03)   & 0.20 / 0.42 (0.01)  & 0.25 / 0.49 (0.05)    \\   
                                                       
    $IR^2$ (Ours)  & \textbf{12.6 (1.7)}   & 13.2 (1.7)  & \textbf{16.4 (2.4)} 
                   & \textbf{8.4 (1.2)}    & \textbf{8.8 (1.2)}   & \textbf{11.4 (2.1)} 
                   & 4.7 (3.6)    & 6.8 (5.0)   & 7.1 (2.0) 
                   & 0.10 / 0.28 (0.08)  & 0.17 / 0.48 (0.20) & 0.21 / 0.61 (0.08)  \\
    \midrule
    & \multicolumn{12}{c}{\textbf{150m $\times$ 150m Forest environment with 2, 4, 6 robots}} \\
    \midrule\midrule
    Preplanned ~\cite{lauren_explore}   
                     & 37.7 (10.2) & 62.4 (12.8)  & 80.4 (4.8) 
                     & 28.8 (8.1)  & 50.7 (9.5)   & 62.5 (1.1) 
                     & 4.6 (3.1)   & 5.1 (2.2)    & 7.9 (0.5) 
                     & 0.07 / \textbf{0.07 (0.01)} & 0.12 / \textbf{0.12 (0.00)}  & 0.23 / \textbf{0.23 (0.02)} \\
    Pursuit ~\cite{mtare_planner}          
                   & \textbf{88.1 (9.5)}  & 103.0 (15.1)  & 110.3 (47.6)
                   & \textbf{46.7 (4.6)}  & 58.2 (6.0)    & 70.9 (17.6) 
                   & 2.3 (3.0)   & 8.6 (2.6)     & 7.5 (2.2) 
                   & 0.22 / 0.58 (0.06)  & 0.28 / 0.63 (0.11)  & 0.31 / 0.64 (0.31) \\
    $IR^2$ (Ours)   & 78.0 (20.1) & \textbf{131.1 (45.6)}  & \textbf{132.6 (38.2)} 
                    & 44.4 (11.4) & \textbf{77.1 (27.6)}   & \textbf{79.4 (23.6)} 
                    & \textbf{2.2 (1.7)}   & \textbf{4.1 (1.5)}     & \textbf{4.0 (1.2)} 
                    & 0.23 / 0.55 (0.15) & 0.21 / 0.59 (0.20)   & 0.16 / 0.59 (0.10) \\
    \midrule
    & \multicolumn{12}{c}{\textbf{340m $\times$ 340m Campus environment with 2, 4, 6 robots}} \\
    \midrule\midrule
    Pursuit ~\cite{mtare_planner}  
    & \textbf{57.7 (15.3)}  & \textbf{59.7 (9.9)} & \textbf{63.5 (7.7)} 
             & \textbf{31.0 (6.1)}   & 31.6 (4.9)  & 35.0 (4.2) 
             & 5.1 (6.3)  & 11.9 (12.0) &  6.1  (3.3) 
             & 0.18 / \textbf{0.50 (0.02)}  & 0.25 / \textbf{0.64 (0.03)} & 0.30 / \textbf{0.73 (0.10)} \\
    $IR^2$ (Ours)  & 46.7 (6.8)  & 57.3 (10.9) & 58.4 (8.6)
                   & 26.8 (3.8)  & \textbf{35.2 (6.3)}  & \textbf{37.3 (3.1)} 
                   & \textbf{3.0 (2.5)}   & \textbf{3.5 (0.8)}   & \textbf{6.1 (2.9)} 
                   & 0.17 / 0.53 (0.10) & 0.33 / 1.17 (0.12) & 0.30 / 1.53 (0.32) \\
    \bottomrule
\end{tabular}

  }
  \label{tab:gazebo_proximity_results}
  \vspace{-0.3cm}
\end{table*}

We compare \textbf{$IR^2$} with a pursuit-based approach~\cite{mtare_planner} (\textit{Pursuit}) and a preplanned-based approach~\cite{lauren_explore} (\textit{Preplanned}). \textit{Pursuit} decomposes the map into exploration volumes defined by frontier clusters and allocates them to robots by solving the Vehicle Routing Problem.
During exploration, robots decide whether and whom to pursue by weighing the distance cost of deviating from its intended route to obtain map information from others, versus staying on-route to explore the area. On the other hand, \textit{Preplanned} involves robots agreeing on and adhering to a specified exploration time budget and rendezvous location. For \textit{Preplanned}, the time budget is pre-set and constant, while the rendezvous position is determined dynamically at a location that minimizes the weighted travel distance for all robots. Note that the original paper~\cite{lauren_explore} includes other robot states intended for task completion, which we leave out for fair comparison. We introduce a 10$s$ gap between launching each robot to encourage distribution, except for \textit{Preplanned} which is set at 1$s$ since it requires all robots to remain in communication range at the beginning.

We calibrate data collection to account for failures observed for each planner, and then report the mean and standard deviation across three runs in Table~\ref{tab:gazebo_proximity_results}. For \textit{Preplanned}, frontier centers are often located in inaccessible regions (e.g. behind fencelines in \textit{Campus}). As such, we do not run \textit{Preplanned} in \textit{Campus}. For \textit{Pursuit}, we notice robots often terminate exploration before achieving 99\% exploration. To account for this, we only select data from runs where \textit{Pursuit} achieves 99\% exploration. For \textbf{$IR^2$}, we notice robots occasionally display faulty distribution at the beginning of exploration, unseen in our simplified training/testing environments. Where relevant, we include one set of such data in every set of three \textbf{$IR^2$} runs for fairness. We believe this problem exists because robots are trained with holonomic constraints, whereas robots in this simulation possess non-holonomic constraints. 
We seek to improve on this \textit{sim-to-real} gap in future works.

We evaluate all methods across the three environments and report the time efficiency, distance efficiency, map area standard deviation, and computation time. Time and distance efficiency are the total volume of map explored per unit time and distance respectively, averaged across all robots. Map area standard deviation measures how equally distributed the exploration task is among the robots. Computation time is the time taken for robots to perform both map post-processing and path planning, averaged across all robots.

\subsubsection{\textbf{Exploration Efficiency}}

In general, we observe an upward trend in both distance and time efficiency when increasing the number of robots. For distance efficiency, we notice that $IR^2$ outperforms \textit{Preplanned} and \textit{Pursuit} in all environments, by at least 27.0\% and 6.6\% respectively, except for the 2-robot \textit{Forest} and \textit{Campus} tests. We believe that these exceptions are due to the tendency of our trained robots to disperse early and rendezvous much later, which works well in our maze-like training environments with many dead-ends. However, in open spaces with many frontiers like \textit{Forest} and the central part of \textit{Campus}, robots may have more difficulty finding each other during rendezvous as their belief of other robots' locations are likely to be very outdated after prolonged dispersion. Nevertheless, we notice this problem ceases when there are more robots, since each robot has less area to explore and thus tends to rendezvous earlier. 

In addition, we observe that performance in distance efficiency may not always translate fully to time efficiency.
This applies to some cases where $IR^2$ is outperformed by \textit{Pursuit} in time efficiency but not distance efficiency, such as in the \textit{Campus} environment.
For \textit{Preplanned}, time efficiency is significantly degraded because of the use of the \textit{MATLAB-ROS toolbox} to interface between \textit{Preplanned} (in MATLAB) and our simulator which introduces computational overhead.

\subsubsection{\textbf{Collaboration Metrics}}

We observe that $IR^2$ achieves the lowest map standard deviation for all experiments in \textit{Forest} and \textit{Campus}. Coupled with $IR^2$'s high distance efficiency, this indicates $IR^2$'s ability to equally and effectively share the exploration task among robots. However, $IR^2$ performs poorly for standard deviation in \textit{Indoor}, because $IR^2$ occasionally exhibits oscillatory behaviors in critical maze-like junctions where robots think others will likely meet them.

\subsubsection{\textbf{Computation Time}}

We observe that \textit{Preplanned} achieves the best computation time across all environments. This is because \textit{Preplanned} performs planning towards frontiers using a map directly, instead of building a graph representation like $IR^2$ or \textit{Pursuit}. However, this comes at the cost of exploration efficiency. We also notice \textit{Pursuit} generally outperforms $IR^2$ in computation time, although $IR^2$ maintains comparable planning speed. This is likely due to our hierarchical graph formulation that maintains a sparse global graph representation of the map for efficient planning. Moreover, it is difficult to fairly compare \textit{Pursuit} in \textit{C++} with $IR^2$ in \textit{Python} (often 1-2 orders of magnitude slower).

\subsection{Additional Studies}

\subsubsection{\textbf{Ablation Studies}}

We validate the importance of our map-surplus utility observation ($s_{i,j}$) and reward ($r_s$) using an ablation study on our curriculum learning framework. We train a separate model without these components. We validate both trained models on three test sets, each containing 100 \textit{Corridor}, \textit{Hybrid}, and \textit{Complex} maps never seen during training. We evaluate its performance using success rate $S(\%)$, simulation steps taken, and distance traveled $D(m)$. To evaluate success, robots need to explore 99\% of the training maps within 196 steps for \textit{Corridor} and \textit{Hybrid} maps (4 robots), and 384 steps for \textit{Complex} maps (5 robots).

Table~\ref{tab:ros_experiments} shows that the model trained with map-surplus utility observation and rewards outperforms the ablated model in all environments and curriculum stages. In addition, curriculum learning improves the success rate in \textit{Complex} maps by 52.0\%, number of steps by 41.1\%, and distance traveled by 20.8\%. However, we notice degradation in model performance particularly for \textit{Corridor} maps after completing stage 2 of the curriculum. This is likely due to the advanced rendezvous strategies learned for \textit{Complex} maps such as waiting at critical junctions to meet up with other robots. Such skills are not as effective in \textit{Corridor} maps that require more basic dispersal and frontier-following strategies.

\subsubsection{\textbf{Signal Strength Communications}}

We validate the versatility of $IR^2$ on signal strength as the communication modality, by performing the same experimental setup as for the proximity model (Table~\ref{tab:gazebo_signal_strength_results}). When compared to the proximity model, the signal strength model performs worse in terms of time and distance efficiency in the \textit{Indoor} environment by at most 29.4\% and 29.8\% respectively. This can be attributed to the absence of information exchange through walls due to the significant decay in signal strength through such a medium. However, our signal strength model performs better in terms of time and distance efficiency in the \textit{Forest} environment by at most 46.4\% and 56.5\% respectively, likely because robots can connect across a long distance as trees act as sparse obstacles. Lastly, the signal strength model outperforms the proximity model in the \textit{Campus} environment in most cases. This is because robots can communicate across long distances in the central part of \textit{Campus}, yet unable to communicate across the walkways blocked by buildings. 

\begin{table}[t]
\centering
% \captionsetup{font={small,stretch=1}}
\captionsetup{font=small}
\caption{Training performance and ablation test.}
\scriptsize 
\begin{tabular}{@{}ccccc@{}}
\toprule
\textbf{Stage} & \textbf{Criteria} & \textbf{\textit{Corridor}} & \textbf{\textit{Hybrid}} & \textbf{\textit{Complex}} \\ \midrule
\multicolumn{5}{c}{Without Map-Surplus Utility ($s_{i,j}$, $r_s$)} \\ \midrule
   & $S(\%)$ & 90.0 & 100.0 & 27.0 \\  
1  & Steps & 95.5 & 49.7  & 344.8  \\
   & $D(m)$ & 1307 (\(\pm 418\)) & 689 (\(\pm 175\)) & 4995 (\(\pm 1216\)) \\
   \midrule
   & $S(\%)$ & 79.0 & 100.0 & 84.0 \\  
2  & Steps & 130.5 & 60.4  & 213.3  \\
   & $D(m)$ & 1589 (\(\pm 449\)) & 725 (\(\pm 174\)) & 3718 (\(\pm 1172\)) \\
   \midrule
\multicolumn{5}{c}{With Map-Surplus Utility ($s_{i,j}$, $r_s$)} \\ \midrule
   & $S(\%)$ & \textbf{97.0} & 100.0 & 35.0 \\  
1  & Steps & \textbf{75.1} & 46.4  & 315.2  \\
   & $D(m)$ & \textbf{1002} (\(\pm 258\)) & 607 (\(\pm 148\)) & 3652 (\(\pm 1480\)) \\
   \midrule
   & $S(\%)$ & 94.0 & \textbf{100.0} & \textbf{87.0} \\  
2  & Steps & 91.0 & \textbf{48.8}  & \textbf{185.8}  \\
   & $D(m)$ & 1212 (\(\pm 424\)) & \textbf{581} (\(\pm 128\)) & \textbf{2892} (\(\pm 1128\)) \\
   \midrule
  \bottomrule
\end{tabular}
\label{tab:ros_experiments}
\end{table}

\begin{table}[t]
  \centering
  \caption{Experimental results (signal strength).}
  \label{tab:results}
  \scalebox{0.72}{
    \begin{tabular}{l||ccc||ccc}
    \toprule
    \textbf{Model} & 
    \multicolumn{3}{c}{Time Efficiency \textbf{$\eta_T$ ($m^3/s$)} $\uparrow$} &
    \multicolumn{3}{c}{Distance Efficiency \textbf{$\eta_D$ ($m^3/m$)} $\uparrow$} \\
    \midrule
    \cmidrule{2-7}

    & \multicolumn{6}{c}{\textbf{2, 3, 4 robots (Indoor); 2,4,6, robots (Forest)}} \\
    
    \midrule\midrule                    
    Indoor        & 8.9 (1.2)   & 12.5 (1.9)   & 12.8 (3.0)  
                  & 5.9 (0.7)   & 8.6 (1.3)    & 8.7 (1.8)    \\ 
    Forest        & 100.3 (20.1)  & 143.7 (4.5)  & 194.1 (28.7)  
                  & 56.6 (10.5)   & 84.6 (4.9)    & 124.3 (23.4)   \\ 
    Campus        & 41.8 (5.5)   & 59.6 (1.1)   & 60.5 (4.4)  
                  & 24.1 (3.2)   & 35.6 (0.9)   & 37.5 (2.4)    \\ 
    \bottomrule
\end{tabular}
  }
  \label{tab:gazebo_signal_strength_results}
  \vspace{-14pt}
\end{table}

\subsection{Experimental Validation}
\begin{figure}[t]
    \centering
    \includegraphics[width=0.85\linewidth]{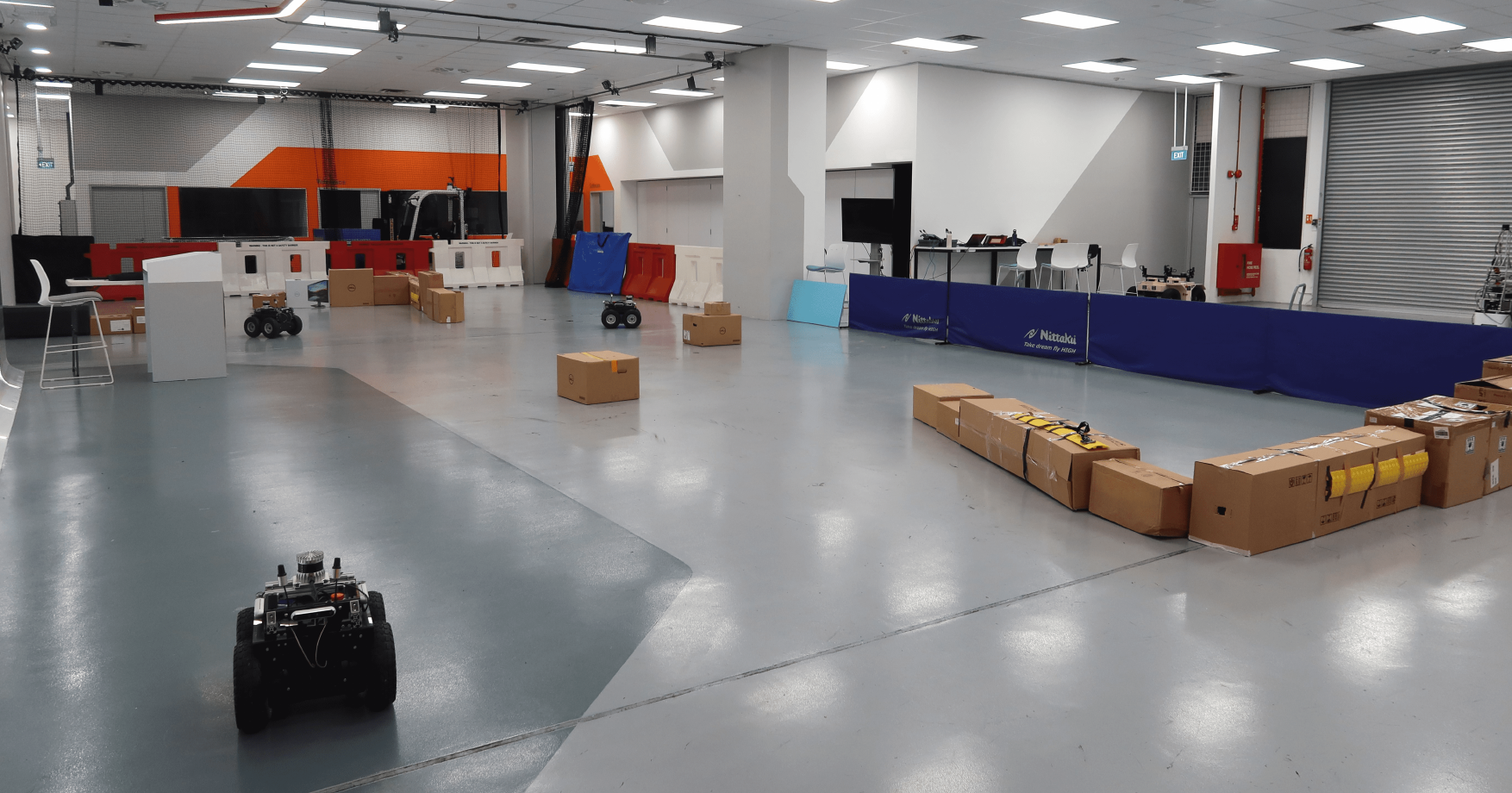}
    \vspace{-0.1cm}
    \caption{Illustration of real-world experimental setup.}
    \label{fig:lab_setup}
    \vspace{-5pt}
\end{figure}

\begin{figure}[t]
    \centering
    \includegraphics[width=1.0\linewidth]{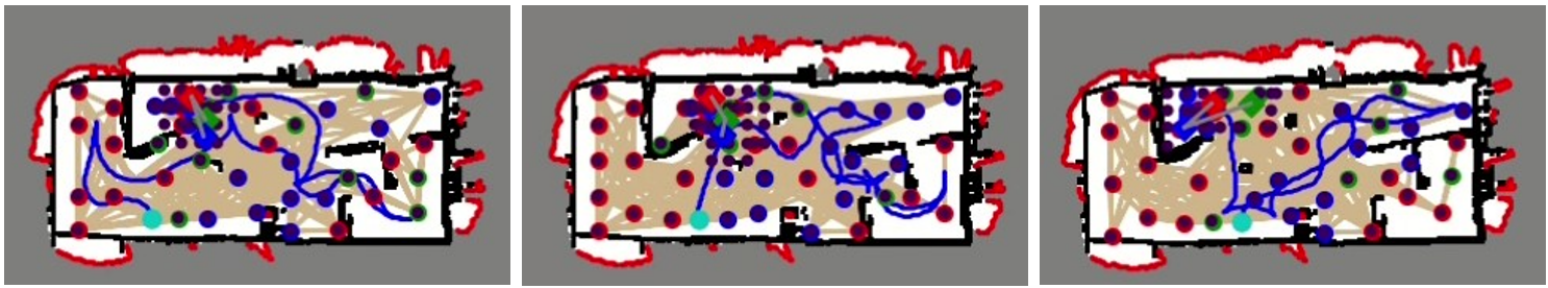}
    \vspace{-0.5cm}
    \caption{Snapshot of the final merged graph representations towards the end of an exploration mission for robots 1 (red), 2 (green), and 3 (blue), from left to right. The colored circles represent each of the robot's graph nodes merged from other robots, the colored diamonds their belief positions, and blue trajectories their paths taken.}
    \label{fig:real_merged_graphs}
    \vspace{-10pt}
\end{figure}

We conduct real-world experiments using three four-wheel differential-drive robots in a $\text{25}m \times \text{10}m$ obstacle-rich environment (Fig.~\ref{fig:lab_setup}). Each robot is equipped with an Ouster 32-plane mapping lidar (up to $3.5$m), and with a Doodlelab mesh radio that allows for information hopping using the proximity-based communication model (up to $3.0$m). The robots executes their own decentralized policy using the same trained model used in Sec.~\ref{training_details}. Throughout the exploration mission, robots repeatedly disconnect and reconnect in an intelligent manner. A snapshot of each robot's final merged sparse graph representations can be seen in Fig.~\ref{fig:real_merged_graphs}. 

\section{CONCLUSION} 
\label{sec:conclusion}

In this paper, we present $IR^2$, a deep reinforcement learning approach to information sharing for multi-robot exploration, particularly in large-scale environments with sparse and intermittent connectivity. $IR^2$ leverages attention-based neural networks and curriculum learning to enable robots to effectively balance the longer-term trade-offs between disconnecting for solo exploration and reconnecting for information sharing. We introduce a novel hierarchical graph formulation to achieve scalability to large-scale environments. Our experiments demonstrate the superiority of our approach against state-of-the-art preplanned and pursuit-based planners, especially in distance efficiency and map area standard deviation. This indicates strong collaboration to achieve high performance and equitable sharing of the exploration task. 

Future research will include studying the impact of more realistic communication models on planner performance by incorporating latency or data packet loss. In addition, while team-based preplanned rendezvous may be inefficient, we hope to explore subteam-based preplanned rendezvous. This means robots will also consider who should be part of the rendezvous agreement, to promote consistent information sharing while avoiding excessive backtracking. Finally, we hope to extend our 2D planner to 3D, in order to benchmark our work in more complicated and realistic environments.

\section*{ACKNOWLEDGMENT}
This work was supported by Singapore Technologies Engineering Ltd, under the Economic Development Board - Industrial Postgraduate Program (Project No. 2022-2130).

\bibliography{ref}

\end{document}